\documentclass[letterpaper, 10 pt, conference]{ieeeconf}  

\usepackage{amsmath}
\usepackage{amsthm}
\usepackage{graphicx}
\usepackage{caption}
\usepackage{subcaption}
\usepackage{amssymb}
\usepackage{array}
\usepackage{url}
\usepackage{color}
\usepackage{nccmath}
\usepackage[font=footnotesize,skip=2pt]{caption}
\usepackage[ruled,vlined]{algorithm2e}

\newcolumntype{C}[1]{>{\centering\arraybackslash}m{#1}}

\IEEEoverridecommandlockouts                              

\usepackage[letterpaper, left=0.667in, right=0.667in, bottom=0.597in, top=0.792in]{geometry}

\usepackage[
    style=ieee,
    doi=false,
    isbn=false,
    url=false,
    eprint=false,
    backend=bibtex,
    natbib=true
    ]{biblatex}

\bibliography{bibliography}

\newcommand{\planneracronym}{STORMMAP}
\newcommand{\plannername}{Stability and Task Oriented Receding-Horizon Motion and Manipulation Autonomous Planner}

\newcommand{\norm}[1]{\left\lVert#1\right\rVert}

\newtheorem{defn}{Definition}

\newtheorem{assum}[defn]{Assumption}

\title{\LARGE \bf
Generating Continuous Motion and Force Plans in Real-Time for Legged Mobile Manipulation
}

\author{Parker Ewen$^{1}$, Jean-Pierre Sleiman$^{2}$, Yuxin Chen$^{1}$, Wei-Chun Lu$^{1}$, Marco Hutter$^{2}$, Ram Vasudevan$^{1}$
\thanks{$^{1}$Authors are with Robotics and Optimization for Analysis of Human Motion Lab, University of Michigan, USA}%
\thanks{$^{2}$Authors are with the Robotic Systems Laboratory, ETH Zürich, Zürich 8092, Switzerland}%
\thanks{This work is supported by the Ford Motor Company via the Ford-UM Alliance under award N022977 and by the Office of Naval Research under Award Number N00014-18-1-2575.}
}

\begin{document}

\maketitle
\thispagestyle{empty}
\pagestyle{empty}

\begin{abstract}

Manipulators can be added to legged robots, allowing them to interact with and change their environment.
Legged mobile manipulation planners must consider how contact forces generated by these manipulators affect the system.
Current planning strategies either treat these forces as immutable during planning or are unable to optimize over these contact forces while operating in real-time.
This paper presents the \plannername{} (\planneracronym{}) that is able to generate continuous plans for the robot's motion and manipulation force trajectories that ensure dynamic feasibility and stability of the platform, and incentivizes accomplishing manipulation and motion tasks specified by a user.
\planneracronym{} uses a nonlinear optimization problem to compute these plans and is able to run in real-time by assuming contact locations are given a-priori, either by a user or an external algorithm.
A variety of simulated experiments on a quadruped with a manipulator mounted to its torso demonstrate the versatility of \planneracronym{}.
In contrast to existing state of the art methods, the approach described in this paper generates continuous plans in under ten milliseconds, an order of magnitude faster than previous strategies.
\end{abstract}

\section{INTRODUCTION} \label{introduction}
Legged robots offer unique advantages when compared to wheeled or aerial robots when autonomously performing tasks in complex terrains while accommodating non-trivial payloads.
To improve the utility of legged robots, additional appendages, such as arms, can be integrated into the platform, thereby providing manipulation capabilities. 
The addition of manipulators can enable legged platforms to interact with and alter their environments; for example, to clear obstacles which may be otherwise impassable.

In fact, manipulation capabilities have been implemented on a number of existing legged platforms.
The torque-controllable quadruped Anymal \cite{hutter2016anymal} was outfitted with a Kinova Jaco six DoF manipulator \cite{campeau2019kinova} to accomplish a variety of tasks \cite{bellicoso2019alma}. 
The hydraulically-actuated quadruped (HyQ) robot \cite{semini2011design}\cite{rehman2016towards} and the Boston Dynamics' SpotMini also have manipulation capabilities. 
For humanoid robots such as Digit \cite{hurst2019walk}, Toro \cite{englsberger2014toro}, and Boston Dynamic's Atlas, arms are an inherent part of their humanoid physiology.

\begin{figure} [!tb]
    \centering
    \includegraphics[width=\columnwidth]{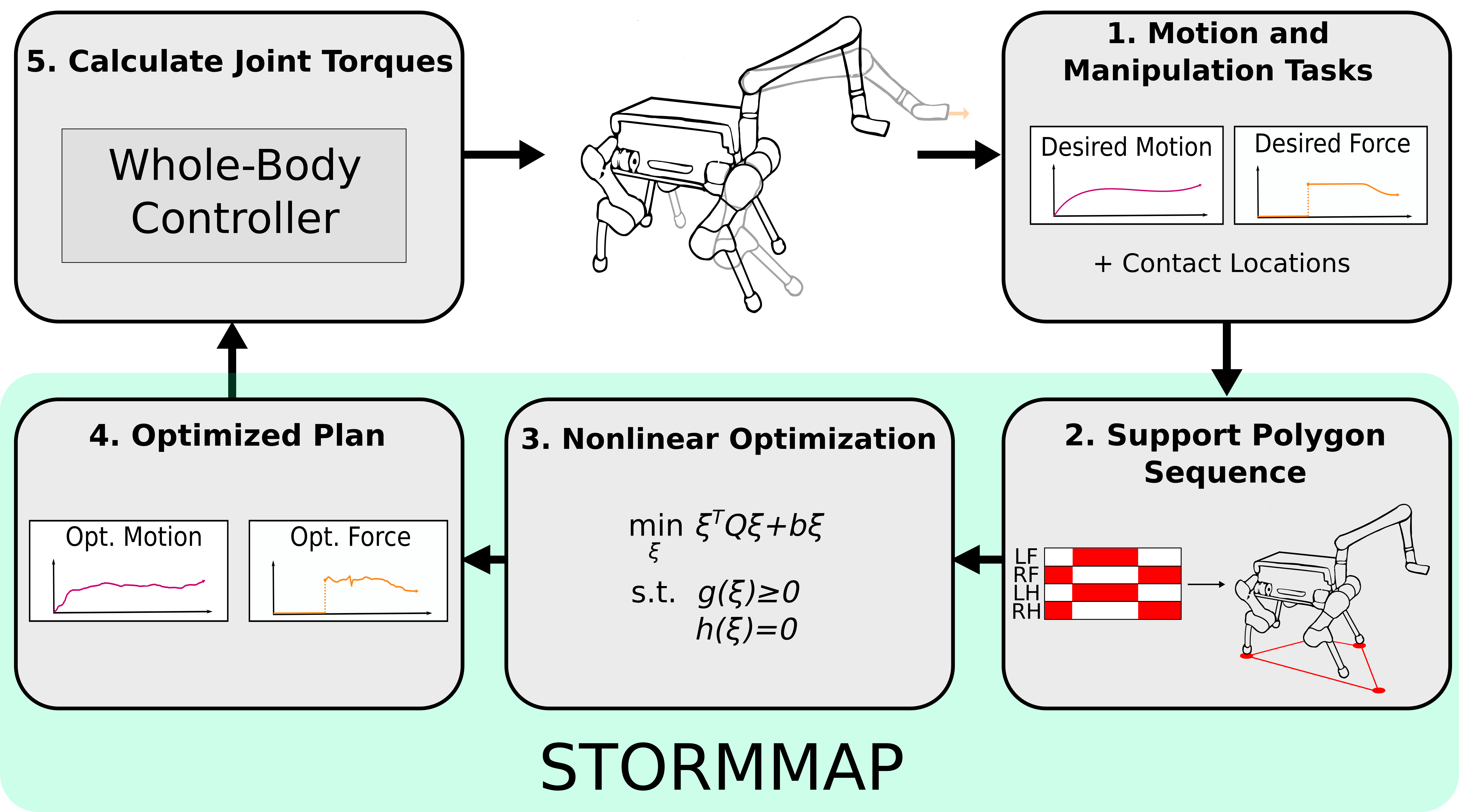}
    \caption{
    Block diagram for the real-time dynamic legged mobile manipulation planner \planneracronym{}. (1) The user specifies desired motion and manipulation tasks for the robot to complete. (2) A sequence of support polygons is generated for one gait cycle to be used for the ZMP stability criterion. (3) A motion trajectory and manipulation force plan is computed using nonlinear optimization. (4) This plan accomplishes the motion and manipulation tasks if it can do so while maintaining dynamic feasibility and stability. (5) The robot's controller then calculates the joint torques and contact forces at the feet to achieve the generated plan.}
    \label{fig:alma_main_figure}
\end{figure}

The addition of manipulation appendages introduces additional forces that must be accounted for during planning.
For instance, if these interaction forces are unaccounted for while assessing the stability of the platform, the forces that enable manipulation may unduly contribute to instability and subsequent failure.
To take full advantage of these additional manipulation capabilities, planning algorithms for these systems must, at a minimum, consider how these new appendages affect the system.
To improve the capabilities of these systems compared to existing state-of-the-art legged mobile manipulation planners, this paper presents a method to jointly plan for the motion trajectory and manipulation force and is able to run in real-time.

The existing literature in planning for legged robotic systems with manipulators can be divided into two categories. 
The first set of approaches, which we refer to as static techniques, are focused on planning the platform's motion while assuming that the forces at the end-effector are unalterable during planning.
The second set of approaches, which we refer to as dynamic techniques, focus on leveraging the additional appendages to improve the stability of the overall system by solving for both the contact locations and contact forces of these new appendages.

Static planning does not allow the forces at the manipulator's end-effector to be altered during planning.
To ensure stability of a plan, these manipulation forces are incorporated into a stability criterion that is typically described using the zero-moment point (ZMP) or centroidal dynamics.
The ZMP stability criterion \cite{vukobratovic1972stability,sardain2004forces} requires that the ZMP remain within the support polygon generated by the robot's feet in contact with the ground and can be extended to include the additional manipulation forces and their effect on the position of the ZMP \cite{takubo2005pushing,stephens2010dynamicforce}. 
Other methods treat the new contact created by the manipulator as an additional point affecting the structure of the support polygon \cite{inomata20103dzmp,harada2003zmp,harada2003gzmp}.
When using a more complex model for motion planning, such as centroidal dynamics, the static forces are considered directly in the equations of motion \cite{dai2014whole}.
By treating the manipulator end-effector forces as static, these strategies rely entirely on the motion of the center-of-mass and contact forces at the feet to maintain stability. 
This can result in the planner either generating a motion plan that maintains stability while sacrificing dynamic feasibility or being unable to find a feasible plan while satisfying the imposed constraints.

In contrast, dynamic approaches solve for both the contact location and force of the end-effector for the purpose of keeping the robot stable while it follows a given base trajectory.
Such strategies have been used to construct plans to robustly traverse complex terrain \cite{lin2018offline,lin2019robust} as well as for walking and object manipulation \cite{bouyarmane2012humanoid}.
Similarly, methods for footstep planning on uneven terrain involving mixed-integer convex optimization \cite{deits2014miqcqp} have been adapted to include hand contacts \cite{ponton2016convex}.
By planning for the contact locations and forces alongside the motion trajectories, these dynamic strategies can use the manipulator to stabilize the robot while completing manipulation tasks.
However, dynamic strategies suffer from either long computational times associated with the complex nature of planning for the contact locations, or utilize coarse time discretizations to speed up computations.
This can make it challenging to implement these planners on robots, as the planning horizon is often shorter than the time it takes to plan.
Additionally, disturbances that occur during the long planning time or between distant time samples are unaccounted for in the new plan, leading to plans that may no longer represent the current state of the system.

The contribution of this paper is the introduction of a real-time dynamic planning strategy, the \plannername{} (\planneracronym) for legged mobile manipulation.
Plans are generated for a time horizon which is 80 times longer than their computational time, allowing these plans to account for disturbances and to be generated in a receding-horizon fashion.
\planneracronym{} jointly plans for both the motion trajectory of the robot and the contact force at a manipulator's end-effector given a contact location.
To the best of our knowledge, this is the fastest dynamic planning strategy in the literature based on planning speed by an order of magnitude \cite{dai2014whole, lin2018offline, lin2019robust, bouyarmane2012humanoid, ponton2016convex}.

The paper is organized as follows:
We first introduce the robot model and contextualize the use of a legged mobile manipulation planner in Section \ref{problem}.
In Section \ref{spline_paramererization} we formulate the parameterization used to reduce the computational complexity of the optimization problem.
Details regarding the nonlinear optimization problem used for \planneracronym{} are laid out in Section \ref{optimization}.
We demonstrate the versatility of \planneracronym{} on the Anymal platform \cite{hutter2016anymal} which is equipped with a Kinova Jaco 6 DoF arm \cite{campeau2019kinova} at its front. 
The planner is validated in a variety of experiments described in Section \ref{planner_behaviours}, which consist of both manipulation tasks and instability-inducing scenarios. 
We include a baseline motion planner for comparison that does not consider the manipulation forces when planning.
Finally, we offer concluding remarks in Section \ref{conclusions}.

\textit{Notation} - The following notation is used throughout the paper: Vectors and vector-valued functions are lowercase and bolded, while sets and matrices are uppercase. 
Vectors are considered as columns. 
$\boldsymbol{f}(t)$ denotes the value that a time-varying function takes at time $t$. 
Subscripts indicate a description and superscripts in parentheses indicate an index. 
An $n\times n$ matrix with diagonal elements $d_1, d_2, ..., d_n$ is given as $\mathrm{diag}(d_1, d_2,..., d_n)$.
For some state $x$, the first and second derivatives w.r.t. time are $\dot{x}$ and $\ddot{x}$ respectively.
An $n\times m$ matrix of zeros is $\boldsymbol{0}_{n\times m}$.
The $L2$-norm is denoted as $\norm{\cdot}_2$.

\section{PROBLEM SETUP} \label{problem}
This section introduces the model of the robot and motivates the use of a legged mobile manipulation planner to generate motion trajectories and manipulation force plans.
Let the generalized coordinates, $\boldsymbol{q}$, be a vector of continuous functions of time that describe the configuration of a robot, $\boldsymbol{u}$ be a vector of differentiable functions of time, and $\dot{\boldsymbol{u}}$ be their time derivatives, each with the following forms:
 
 \begin{equation}
     \boldsymbol{q}=
     \begin{bmatrix}\boldsymbol{q}_{b_P} \\ \boldsymbol{q}_{b_R} \\ \boldsymbol{q}_j \end{bmatrix}
     \;\;
     \boldsymbol{u}=
     \begin{bmatrix}\boldsymbol{v}_b \\ \boldsymbol{\omega}_{b} \\ \dot{\boldsymbol{q}}_j\end{bmatrix}
     \;\;
     \dot{\boldsymbol{u}}=
     \begin{bmatrix}\boldsymbol{a}_b \\ \dot{\boldsymbol{\omega}}_{b} \\ \ddot{\boldsymbol{q}}_j\end{bmatrix}
 \end{equation}
 
\noindent $\boldsymbol{q}_{b_P}$ and $\boldsymbol{q}_{b_R}$ are functions of time describing the robot's center-of-mass (CoM) position and rotation, respectively, $\boldsymbol{v}_b$ and $\boldsymbol{\omega}_{b}$ are functions of time describing the robot's linear and angular velocity, respectively, $\boldsymbol{a}_b$ and $\dot{\boldsymbol{\omega}}_b$ are functions of time describing the robot's linear and angular accelerations, respectively, and $\boldsymbol{q}_j$ are functions of time describing the $n_j$ actuated joint positions of the robot. 
We assume that position of the robot, $\boldsymbol{q}_{b_P}$, is parameterized using Cartesian coordinates.

The high-fidelity model for a legged robot with a manipulator derived using the projected Newton-Euler equation with the constraint compliant Lagrange formulation \cite[Eq. 4.55a]{hahn2013rigid} can be written as follows:
\begin{equation} \label{equation_of_motion}
    M(\boldsymbol{q}(t))\dot{\boldsymbol{u}}(t) + \boldsymbol{b}(\boldsymbol{q}(t), \boldsymbol{u}(t)) = S^T \boldsymbol{\tau}(t) + J(t)^T \boldsymbol{f}(t)
\end{equation}
where at time $t$, $M(\boldsymbol{q}(t))$ is the mass matrix, $b(\boldsymbol{q}(t), \boldsymbol{u}(t))$ is the vector of Coriolis, centrifugal and gravity terms,  $\boldsymbol{\tau}(t)$ are the torques acting in the direction of the generalized coordinates, $S$ is the selection matrix for the actuated joints, $\boldsymbol{f}(t)$ are the contact forces acting on the robot, and $J(t)$ is the contact Jacobian matrix.
We make the following assumption about this model:
\begin{assum}{} \label{assum:controller}
    There exists a controller which uses \eqref{equation_of_motion} to calculate the joint torques and contact forces for the robot to achieve some desired base acceleration, $\boldsymbol{a}_b$, and a desired manipulation force, $\boldsymbol{f}_m$, that is a subset of the set of contact forces, $\boldsymbol{f}$. 
\end{assum}
\noindent Various techniques have been proposed in the literature to construct a controller that satisfies this assumption \cite{ hutter2016anymal,bellicoso2019alma,rehman2016towards,stephens2010dynamicforce,ponton2016convex,apgar2018fast,Mastalli_2020}.

The goal of our planner is to generate trajectories for $\boldsymbol{q}_{b_P}$, $\boldsymbol{v}_b$, $\boldsymbol{a}_b$, and $\boldsymbol{f}_m$ over a time interval $[0,T]$ that are dynamically feasible, maintain stability, and can accomplish a pre-defined motion or manipulation task, if possible. 
We formulate this planner using the following nonlinear optimization problem:
\begin{subequations} \label{original_opt}
    \begin{align}
    \min_{ \boldsymbol{q}_{b_P}, \boldsymbol{v}_b, \boldsymbol{a}_b, {\boldsymbol{f}}_{m}} \quad & \mathrm{Cost}( \boldsymbol{q}_{b_P}, \boldsymbol{v}_b, \boldsymbol{a}_b,  \boldsymbol{f}_{m}) \label{cost}\\
    \textrm{s.t.} \quad & \mathrm{Motion \ Spline \ Junctions}( \boldsymbol{q}_{b_P}, \boldsymbol{v}_b) \label{eq:motion_junctions}\\
                        & \mathrm{Initial \ Spline \ Point}( \boldsymbol{q}_{b_P}, \boldsymbol{v}_b) \label{eq:initial_point}\\
                        & \mathrm{Friction \ Pyramid}(\boldsymbol{a}_b,  \boldsymbol{f}_{m}) \label{eq:friction_pyramid}\\
                        & \mathrm{ZMP \ Stability}( \boldsymbol{q}_{b_P}, \boldsymbol{a}_b,  \boldsymbol{f}_{m}) \label{eq:zmp_stability}\\
                        & \mathrm{Force \ Limits}( \boldsymbol{f}_{m}) \label{eq:force_limits}\\
                        & \mathrm{Free \ Motion \ Direction}( \boldsymbol{f}_{m}) \label{eq:free_motion}
    \end{align}
\end{subequations}
\noindent We achieve a smooth motion trajectory by constraining the position and velocity of the robot to be continuous \eqref{eq:motion_junctions} and constrain the initial position and velocity of the optimized trajectory to be equivalent to the measured value at the start of the optimization \eqref{eq:initial_point}.
The friction pyramid models the relationship between the forces at the contact locations and the accelerations of the robot to ensure dynamic feasibility of the generated plans \eqref{eq:friction_pyramid}.
The stability of the platform is maintained in the presence of manipulation forces using the ZMP stability model \cite{vukobratovic1972stability} that includes these forces \eqref{eq:zmp_stability}.
Lastly, the manipulation forces are kept within the limits of the arm \eqref{eq:force_limits} and are also prohibited in free motion directions \eqref{eq:free_motion}.
The cost is used to incentivize the completion of tasks.

In practice, this optimization problem is solved by discretizing the functions that appear as decision variables at pre-specified sampling times.
The accuracy of the approximation is dependent on the chosen time discretization, with a finer discretization leading to a more accurate approximation but more parameters, thereby increasing the solution space of the optimization and the computational time required to solve the problem.
We overcome this curse of dimensionality by following the example set in \cite{hutter2016anymal, bellicoso2019alma} and parameterize the plan using quintic splines.
We reformulate \eqref{original_opt} using the spline parameterization which results in a nonlinear optimization problem whose average computational time is under 10 milliseconds.
The speed of the optimization is such that we can use \planneracronym{} to plan in a receding horizon fashion as the planning horizon is longer than the time it takes to compute a plan.
 
\section{SPLINE PARAMETERIZATION} \label{spline_paramererization}
This section describes the parameterization used to encode the motion and manipulation force trajectories as quintic splines. 
The time-dependent CoM trajectory can be parameterized using a set of quintic spline parameters.
Spline parameters for the motion trajectory are denoted by $\alpha$ while those for manipulation force are denoted by $\beta$.
For instance, the spline parameterization for the \textit{x} direction of the motion trajectory is 
\begin{equation} \label{splines}
    x(t) = \boldsymbol{t}^\intercal(t) \boldsymbol{\alpha}_x, \;\;\;
    \dot{x}(t) = \dot{\boldsymbol{t}}^\intercal(t) \boldsymbol{\alpha}_x, \;\;\; 
    \ddot{x}(t) = \ddot{\boldsymbol{t}}^\intercal(t) \boldsymbol{\alpha}_x, 
\end{equation}
where $\boldsymbol{\alpha}_x = [\alpha_{x,5}, \alpha_{x,4},\alpha_{x,3}, \alpha_{x,2}, \alpha_{x,1}, \alpha_{x,0}]^\intercal$ and $\boldsymbol{t}=[t^5, t^4, t^3, t^2, t, 1]^\intercal$.
The same parameterization is used for the manipulation force at the end effector.
We can describe the full motion or manipulation force trajectory as:
\begin{equation}
\label{eq:spline_motion_trajectory}
    \begin{bmatrix}
    x(t) \\
    y(t) \\
    z(t)
    \end{bmatrix}
    = T(t) \begin{bmatrix}
    \boldsymbol{\alpha}_x \\
    \boldsymbol{\alpha}_y \\
    \boldsymbol{\alpha}_z
    \end{bmatrix}
\end{equation}
where $T(t)=\mathrm{diag}(\boldsymbol{t}^\intercal(t),\boldsymbol{t}^\intercal(t),\boldsymbol{t}^\intercal(t))$.

Next, we account for the hybrid dynamics of legged robotic systems.
To account for the contact transitions at the feet during walking, we consider the following:

\begin{assum} \label{assum:contact_sequence}
    Given a user-specified desired velocity, which is assumed constant over the motion trajectory, a set of foot contact locations can be computed using the linear inverted pendulum model \cite{gehring2016footholds}.
    This contact sequence is calculated for one gait cycle with a velocity-dependant time horizon.
    A new support polygon is generated for each lift-off and touch-down event of the feet, and using a dynamic trotting gait, this yields a sequence of five support polygons over a gait cycle.
\end{assum}
\noindent A single piecewise quintic spline is used per support polygon in the generated sequence described in Assumption \ref{assum:contact_sequence}.
This means that for each optimization, we solve for the parameters of five splines associated with the motion trajectory.
These splines are indexed using $i$ and each has a domain $[t^{(i)}_0, t^{(i)}_f]$.
We concatenate the spline parameters for an entire motion trajectory composed of a series of piecewise splines into a single vector $\boldsymbol{\alpha}$.

To expand on Assumption \ref{assum:contact_sequence}, we form a desired motion trajectory given the current position of the robot and the user-specified constant desired velocity. We evaluate this desired motion trajectory over the entire optimization horizon as follows:

\begin{defn} \label{def:motion_task}
    Let $\boldsymbol{\pi}_\alpha$ denote the desired position trajectory of the robot, with $\boldsymbol{\dot{\pi}}_\alpha$ and $\boldsymbol{\ddot{\pi}}_\alpha$ the desired velocity and acceleration trajectories, respectively. We evaluate $\boldsymbol{\pi}_\alpha$ by first setting $\boldsymbol{\pi}_\alpha(0)$ to the measured position of the robot and calculating $\boldsymbol{\pi}_\alpha(T)$ by integrating the constant desired velocity from Assumption \ref{assum:contact_sequence} over the optimization horizon $[0,T]$. The desired velocity trajectory is set to the constant user-specified velocity  and the desired acceleration trajectory is set to zero.
\end{defn}
    
To account for the contact transitions at the manipulator, we make the following assumption:
\begin{assum} \label{assum:force_transition}
    The contact locations and times of contact for the manipulator are known \textit{a priori} to the planner. 
    Contact occurs when the manipulator has at least one direction of constrained motion, and contact is broken when all directions at the manipulator are in free motion. 
\end{assum}
\noindent Note that the contact locations and times of contact for the manipulator can be either specified by the user or by an external algorithm.
When contact occurs or is broken, we call this a \emph{contact transition}.
We denote the number of contact transitions that occur over the optimization horizon as $n_m$.
Similar to the motion trajectory, we construct a sequence of force splines to describe the force across a time horizon. 
We use one force spline to describe the force before a transition, indexed as $j$ with the domain of each spline equal to $[t^{(j)}_0, t^{(j)}_f]$, and one force spline for the force after the transition, indexed by $j+1$. 
This allows for a discontinuity in the force at the transition. 
As in the case of the motion trajectory, we concatenate all of the spline parameters for the manipulation force trajectory into a single vector that we denote by $\boldsymbol{\beta}$.

When possible, we want the generated manipulation force plan to achieve some user-specified manipulation task. To do this, we define the set of \emph{force events} which represent these desired manipulation tasks:

\begin{defn} \label{def:force_task}
    Let a force event $\boldsymbol{\pi}_\beta^{(j)}$ denote the desired manipulation force over a range $[t_0^{(j)}, t_f^{(j)})$ which occurs before the $j^{th}$ contact transition.
    The set of force events is the set of all such force events arranged sequentially, denoted as $\boldsymbol{\pi}_\beta = \{ \boldsymbol{\pi}_\beta^{(j)} | j \in J \}$, such that, combined, they span the range of the optimization horizon $[0, T]$.
    Here, $J \in 1,..., n_m+1$.
\end{defn}

It should also be noted that a force event is user-defined and based on the manipulation task.
The first and second time derivatives for this pre-defined force function are therefore $\boldsymbol{\dot{\pi}}_\beta$ and $\boldsymbol{\ddot{\pi}}_\beta$ respectively.
If a constant force is desired, these are set to zero.

Using the spline parameterization for the motion trajectory and manipulation force, we rewrite the optimization problem in \eqref{original_opt} to optimize over the set of spline parameters.

\section{OPTIMIZATION} \label{optimization}

This section describes the formulation of the cost functions and constraints of the nonlinear optimization problem \eqref{original_opt}. 
The vector of motion and force spline parameters mapping to the state trajectories is the optimization variable.
In practice, it is challenging to validate a constraint over a continuous time interval.
We therefore discretize the time based on the following assumption:
\begin{assum} \label{assum:time_samples}
    Take each piecewise motion and force spline and sample them at six evenly spaced time intervals, including the initial and final point of the spline.
\end{assum}
We denote the set of sample times as $S_t$.
This discretization is needed, as we cannot enforce constraints over an entire trajectory.
The fineness of this discretization makes it unlikely that the generated continuous trajectories will violate a constraint between time samples.
This discretization does not increase the solution space of problem \eqref{original_opt} due the the spline parameterization, even though the same number of constraints are used for both the original and parameterized optimizations.

The remainder of this section details the formulation of the nonlinear optimization problem used to generate motion and manipulation force plans. The costs of the optimization problem in \eqref{original_opt} are described in Section \ref{sec:cost_functions}, while the equality and inequality constraints are formulated in Sections \ref{sec:equality_constraints} and \ref{sec:inequality_constraints}, respectively.

\subsection{Cost} \label{sec:cost_functions}

The cost in the optimization problem \eqref{original_opt} is assumed to be a positive definite quadratic form.
In particular, the quadratic cost,  $Q$, and linear cost, $\boldsymbol{b}$, take the following form:
\begin{equation}\label{eq:generalcost}
\scalebox{.9}{$
\begin{aligned}
    Q &= \mathrm{diag}(Q_{\alpha}^{(1)}+Q_{\alpha_{im}}, \cdots, Q_{\alpha}^{(5)}, Q_{\beta}^{(1)}+Q_{\beta_{im}}, \cdots Q_{\beta}^{(n_m)}) \\
    \boldsymbol{b}^\intercal &= [\boldsymbol{b}_\alpha^{(1)\intercal}+\boldsymbol{b}_{\alpha_{im}}^\intercal, \cdots, \boldsymbol{b}_\alpha^{(5) \intercal}, \boldsymbol{b}_\beta^{(1)\intercal}+\boldsymbol{b}_{\beta_{im}}^{\intercal}, \cdots, \boldsymbol{b}_\beta^{(n_m)\intercal}]
\end{aligned}
$}
\end{equation}
\noindent where $Q_{\alpha}^{(i)} \in \mathbb{R}^{6\times 6}$ and $\boldsymbol{b}^{(i)}_\alpha \in \mathbb{R}^{6\times 1}$ are the quadratic and linear costs for the $i^{th}$ motion spline, and $Q_{\beta}^{(j)} \in \mathbb{R}^{6\times 6}$ and $\boldsymbol{b}_\beta^{(j)} \in \mathbb{R}^{6 \times 1}$ are the quadratic and linear costs for the $j^{th}$ force spline. The costs $Q_{\alpha_{im}}^{(1)}$ and $\boldsymbol{b}_{\alpha_{im}}^{(1)}$ minimize the difference between the initial acceleration of the motion trajectory and the measured acceleration, while $Q_{\beta_{im}}^{(1)}$ and $\boldsymbol{b}_{\beta_{im}}^{(1)}$ minimize the difference between the initial force and measured force.

The cost per motion spline is a linear combination of several time-dependent costs and one time-independent cost.
We compute the quadratic and linear cost terms for each motion spline as:
\begin{equation} \label{total_cost}
\begin{aligned}
    Q_{\alpha}^{(i)} &= Q_{acc}^{(i)} + \sum_{t_s \in S_t} (Q_{\alpha_\pi}^{(i)}(t_s) + Q_{\alpha_d}^{(i)}(t_s)) \\
    \boldsymbol{b}_\alpha^{(i)} &= \sum_{t_s \in S_t} (\boldsymbol{b}_{\alpha_\pi}^{(i)}(t_s) + \boldsymbol{b}_{\alpha_d}^{(i)}(t_s)
\end{aligned}
\end{equation}

\noindent where $Q_{\alpha_\pi}^{(i)}$ is a cost to incentivize a desired task, $Q_{\alpha_d}^{(i)}$ penalizes deviations between subsequent plans and  $Q_{acc}^{(i)}$ penalizes large accelerations over the spline.
An equivalent calculation to \eqref{total_cost} is used for the cost per force spline, with individual costs denoted using $\beta$.
The specific elements of \eqref{total_cost} are more formally defined in the remainder of this subsection.

\subsubsection{Desired Motion and Manipulation Tasks}

We want the generated motion and manipulation force plans to achieve desired tasks, when feasible.
To incentivize these tasks, we use the desired motion trajectory, $\boldsymbol{\pi}_\alpha$, and its derivatives (Definition \ref{def:motion_task}), along with the set of force events, $\boldsymbol{\pi}_\beta$, representing the desired manipulation tasks (Definition \ref{def:force_task}).
We minimize the difference between values sampled from the motion and manipulation force plans and their respective desired trajectories at each sampling time $t_s$ in a least-squares manner.
An example for the motion splines follows, where $i$ is the $i^{th}$ motion spline corresponding to the sampling time:
\begin{equation} \label{path_reg_cost}
\scalebox{.85}{$
    \begin{aligned}
        \boldsymbol{\alpha}^{(i)\intercal}Q_{\alpha_\pi}^{(i)}(t_s)\boldsymbol{\alpha}^{(i)} + \boldsymbol{b}_{\alpha_{\pi}}^{(i)}(t_s)\boldsymbol{\alpha}^{(i)} = &\norm{T(t_s) \boldsymbol{\alpha}^{(i)}- \boldsymbol{\pi}_t(t_s)}_2 + \\
        + &\norm{\dot{T}(t_s) \boldsymbol{\alpha}^{(i)}- \boldsymbol{\dot{\pi}}_t(t_s)}_2  + \\
        + &\norm{\ddot{T}(t_s) \boldsymbol{\alpha}^{(i)}- \boldsymbol{\ddot{\pi}}_t(t_s)}_2
    \end{aligned}
$}
\end{equation}
An equivalent formulation is used to calculate the costs for the $j^{th}$ force spline, $Q_{\beta_\pi}^{(j)}(t_s)$ and $\boldsymbol{b}_{\beta_{\pi}}^{(j)}(t_s)$.

\subsubsection{Deviation from Previous Plans} \label{subsec:prev_plan}

Because the optimization problem we are solving is nonlinear, each time we solve it, we initialize the solver using the solution found in the previous iteration.
Note, we solve the optimization problem using an SQP solver. 
Given $t_d$, the time which has elapsed since the start of the previous plan, and a sampled time, $t_s$, we minimize the difference between the current plan and the previous successfully computed plan using least squares:
\begin{equation} \label{prev_plan_cost}
\scalebox{.75}{$
\begin{aligned}
    \boldsymbol{\alpha}^{(i)\intercal} Q_{\alpha_d}^{(i)}(t_s)\boldsymbol{\alpha}^{(i)} + \boldsymbol{b}_{\alpha_{d}}^{(i)}(t_s)\boldsymbol{\alpha}^{(i)} =&\norm{T(t_s) \boldsymbol{\alpha}^{(i)} - T(t_s+t_d) \boldsymbol{\alpha}^{(i)}_{prev}}_2 +\\
    +&\norm{\dot{T}(t_s) \boldsymbol{\alpha}^{(i)} - \dot{T}(t_s+t_d) \boldsymbol{\alpha}^{(i)}_{prev}}_2 +\\
    +&\norm{\ddot{T}(t_s) \boldsymbol{\alpha}^{(i)} - \ddot{T}(t_s+t_d) \boldsymbol{\alpha}^{(i)}_{prev}}_2 \\
\end{aligned}
$}
\end{equation}
An equivalent formulation exists for the force spline costs $Q_{\beta_d}^{(j)}(t_s)$ and $\boldsymbol{b}_{\beta_{d}}^{(j)}(t_s)$.

\subsubsection{Initial and Measured State Values} \label{subsec:current_measured}

Large differences between the initial acceleration and manipulation force of the optimized plan and measured accelerations, $\boldsymbol{a}^{*}_b$, and manipulation forces, $\boldsymbol{f}_{m}^{*}$, can cause jumps that may be infeasible to achieve.
Minimizing the difference between the initial and measured values can prevent these jumps:
\begin{equation} \label{initial_motion}
\scalebox{.9}{$
    \boldsymbol{\alpha}^{(1)\intercal} Q_{\alpha_{im}} \boldsymbol{\alpha}^{(1)\intercal} + \boldsymbol{b}_{\alpha_{im}}(t_s)\boldsymbol{\alpha}^{(1)} = \norm{\ddot{T}(0) \boldsymbol{\alpha}^{(1)} - \boldsymbol{a}^{*}_b}_2
$} \end{equation}

\begin{equation} \label{initial_force}
\scalebox{.9}{$
     \boldsymbol{\beta}^{(1)\intercal} Q_{\beta_{im}} \boldsymbol{\beta}^{(1)\intercal} + \boldsymbol{b}_{\beta_{im}}(t_s)\boldsymbol{\beta}^{(1)} = \norm{T(0) \boldsymbol{\beta}^{(1)} - \boldsymbol{f}^{*}_m}_2
$}\end{equation}
\noindent These quadratic and linear costs are used in \eqref{eq:generalcost} for the first motion and force spline only.

\subsubsection{Minimize Acceleration} \label{subsec:min_accel}

We want the generated motion trajectory to be as close to the user-specified trajectory, which has a constant velocity, as possible. Thus, we minimize the accelerations of the generated plan.
Using the method adopted from \cite{kalakrishnan2010} and implemented in \cite{bellicoso2017dynamic, bellicoso2018dynamic}, we derive the quadratic cost by squaring and integrating the acceleration over the time duration of each spline and solve for the central term $Q_{acc}^{(i)}=\int_{t^{(i)}_0}^{t^{(i)}_fn}\ddot{\boldsymbol{t}}^\intercal \ddot{\boldsymbol{t}} \;dt$ . 
This implementation minimizes the cost over the entire motion spline.
For a more thorough treatment of this cost, see \cite{bellicoso2017dynamic, bellicoso2018dynamic}. 
Note, this additional cost is only used for the motion splines.


\subsection{Equality Constraints} \label{sec:equality_constraints}
Three equality constraints are used in the optimization problem \eqref{original_opt}.
They ensure the piecewise motion splines are connected and continuous in their first derivative \eqref{eq:motion_junctions}, that the optimized initial position and velocity are equal to the measured values \eqref{eq:initial_point}, and that manipulation forces cannot be applied in a direction of free motion \eqref{eq:free_motion}.

\subsubsection{Motion Spline Junctions}

Recall from Assumption \ref{assum:contact_sequence} that a sequence of piecewise motion splines are used to account for the hybrid nature of the system. To ensure these splines connect and are continuous in the first derivative, we formulate the following equality for \eqref{eq:motion_junctions}:
\begin{equation} \label{spline_junct}
    \begin{bmatrix}
        T(t_{f}^{(i)}) & T(t_0^{(i+1)}) \\
        \dot{T}(t_{f}^{(i)}) & \dot{T}(t_0{}^{(i+1)})
    \end{bmatrix}
    \begin{bmatrix}
       \boldsymbol{\alpha}^{(i)} \\
       \boldsymbol{-\alpha}^{(i+1)}
    \end{bmatrix} = \boldsymbol{0}_{2\times 1}
\end{equation}

\subsubsection{Initial Spline Point}

If the initial position and velocity of the optimized motion trajectory are not equivalent to the measured values, this would cause large initial accelerations during implementation. 
To prevent this, we ensure that the initial point of the first motion spline at time zero is equal to the measured position of the robot, $r^{*}$.
Likewise, we ensure the initial velocity is set to the measured velocity, $v^{*}$, which leads to the constraint \eqref{eq:initial_point}:
\begin{equation}
\begin{bmatrix}
    T(0) \\
    \dot{T}(0)
\end{bmatrix}
     \boldsymbol{\alpha}^{(1)} = 
\begin{bmatrix}
    \boldsymbol{r}^{*} \\
    \boldsymbol{v}^{*}
\end{bmatrix} 
\end{equation}

This constraint is only for the first point of the optimized motion trajectory, and thus this constraint only applies to the first motion spline.

\subsubsection{Free Motion Direction}

To maintain dynamic feasibility, we ensure forces cannot be applied in directions for which there is free motion by setting these forces equal to zero.
This is done by specifying the direction of free motion through a selection matrix $S^{(j)}=\mathrm{diag}(S_x^{(j)}, S_y^{(j)}, S_z^{(j)})$, where \eqref{eq:free_motion} is then:
\begin{equation} \label{freemotion}
    S^{(j)} \boldsymbol{\beta}^{(j)} = \boldsymbol{0}_{6 \times 1},
\end{equation}
where each component $S^{(j)}_{i} \in \mathbb{R}^{6 \times 6}$ is defined as follows:
\begin{equation}
    S^{(j)}_i=
    \begin{cases}
        \boldsymbol{\mathbb{I}}_{6 \times 6} & \text{for free-motion}\\
        \boldsymbol{0}_{6 \times 6} & \text{otherwise}
    \end{cases}
\end{equation}


\subsection{Inequality Constraints} \label{sec:inequality_constraints}
The inequality constraints ensure that the generated motion trajectory and manipulation plans remain dynamically feasible and stable. This is done using the friction pyramid model \eqref{eq:friction_pyramid} and ZMP stability criterion \eqref{eq:zmp_stability}. Additionally, we constrain the magnitude of the manipulation force to lie within the limits of the arm \eqref{eq:force_limits}.

\subsubsection{Friction Pyramid} \label{sec:frict_pyr}

Feasibility of the motion and manipulation plan requires the robot's contact forces can generate the desired accelerations without slipping. 
The friction pyramid model \cite{trinkle1997dynamic} used in \eqref{eq:friction_pyramid}, a linearized approximation of the friction cone model that also includes the forces at the manipulator's end-effector, can describe this condition:
\begin{equation} \label{friction_pyramid}
\scalebox{.88}{$
    \begin{bmatrix}
        1 & 0 & -\mu\\
        -1 & 0 & -\mu\\
        0 & 1 & -\mu \\
        0 & -1 & -\mu
    \end{bmatrix}
    \boldsymbol{a}_b(t_s) + \begin{bmatrix}
        1 & 0 & -\mu \\
        1 & 0 & -\mu \\
        0 & 1 & -\mu \\
        0 & 1 & -\mu
    \end{bmatrix}
    \boldsymbol{f}_m(t_s) \leq \begin{bmatrix}
        \mu m g \\
        \mu m g \\
        \mu m g \\
        \mu m g
    \end{bmatrix}
    $}
\end{equation}
\noindent where $m$ is the whole-body mass of the robot, $g$ is the gravitational constant $-9.81 \frac{m}{s^2}$, and $\mu$ is the coefficient of friction.
Forces at the manipulator can be used to increase the feasible accelerations of the base if the tangential forces at the feet cannot accommodate these accelerations.
This constraint applies to each sample time $t_s \in S_t$.

\subsubsection{ZMP Stability} \label{sec:zmp}

One method of ensuring stability of a legged system is by using the ZMP stability criterion, which simplifies the robot model to that of an inverted pendulum \cite{vukobratovic1972stability}.
This constraint requires that the ZMP remain within the support polygon for all time.
Similar to the method employed by \cite{takubo2005pushing, stephens2010dynamicforce}, we include the manipulation force into the ZMP derivation and evaluate the constraint at each sample time $t_s \in S_t$:
\begin{equation} \label{zmp}
\scalebox{1}{$
    \boldsymbol{r}_{zmp}(t_s) = \frac{\boldsymbol{n}(t_s) \times \boldsymbol{\tau}(t_s)}{\boldsymbol{n}(t_s) \cdot \boldsymbol{f}(t_s)}
$}
\end{equation}
\begin{equation} \label{zmp_force}
\scalebox{.85}{$
    \boldsymbol{f}(t_s) = m(\boldsymbol{g}-\boldsymbol{a}_{b}(t_s)) + \boldsymbol{f}_{m}(t_s)
$}
\end{equation}
\begin{equation} \label{zmp_moment}
\scalebox{.85}{$
\begin{split}
    \boldsymbol{\tau}(t_s) = &\boldsymbol{q}_{b_P}(t_s) \times m(\boldsymbol{g}-\boldsymbol{a}_{b}(t_s)) \ldots \\
    + &(\boldsymbol{q}_{b_P}(t_s) + C_{Ic} \boldsymbol{r}_{c-m}) \times \boldsymbol{f}_{m}(t_s)
\end{split}
$}
\end{equation}
\noindent where, at time $t_s$, $\boldsymbol{n}(t_s)$ is the normal to the ground, $\boldsymbol{f}(t_s)$ the force exerted by the robot onto the environment \eqref{zmp_force}, and $\boldsymbol{\tau}(t_s)$ is the resultant moment \eqref{zmp_moment}, $m$ the robot's whole-body mass, $\boldsymbol{g}=[0,0,g]^\intercal$, $C_{Ic}(t_s)$ the rotation matrix between the inertial frame and the body frame, and $\boldsymbol{r}_{c-m}(t_s)$ the position of the CoM to the manipulation force center in the body frame which is assumed constant over the planning horizon.

Let $\boldsymbol{d}(t_s)=[\boldsymbol{a}(t_s), \boldsymbol{b}(t_s), \boldsymbol{0}_{4\times 1}]$ and $\boldsymbol{c}(t_s)$ represent the vectorized line variables of the support polygon at sample time $t_s$.
The ZMP constraint \eqref{eq:zmp_stability} is satisfied if:

\begin{equation}
    \boldsymbol{d}(t_s) \boldsymbol{r}_{zmp}(t_s) + \boldsymbol{c}(t_s) \leq \boldsymbol{0}_{4\times 1}
\end{equation}
We can see from \eqref{zmp_moment} that this constraint is nonlinear.

\subsubsection{Force Limit} \label{sec:force_lim}

Implementing manipulation force limits serves two purposes: remaining within the actuator limits of the arm and mandating force direction.
The first is accomplished through the relation of the arm torque limits with the end-effector force using the contact Jacobian $J(t_s) \in \mathbb{R}^{d \times 3}$ for an arm with $d$ actuators.
It is assumed that the arm remains within the same configuration for the entirety of the optimization time horizon.
If $\boldsymbol{\tau}_{lim} \in \mathbb{R}^{d\times 1}$ are the actuator limits of the arm, the following limits the force exertion at the end-effector:
\begin{equation}
    J^\intercal \boldsymbol{f}_{m}(t) \leq \boldsymbol{\tau}_{lim}
\end{equation}
Force direction is also specified using hard bounds on the force.
Requiring the force to be either greater than or less than zero dictates whether the manipulation force is pushing or pulling on the environment, respectively.
For each $t_s \in S_t$, one can represent \eqref{eq:force_limits} as:
\begin{equation}
    \boldsymbol{f}_{lim}^{-} \leq \boldsymbol{f}_{m}(t_s) \leq \boldsymbol{f}_{lim}^{+}.
\end{equation}
 
\section{EXPERIMENTS} \label{planner_behaviours}
This section demonstrates the performance of the algorithm proposed in this paper in simulation on the Anymal robot. 
We describe the implementation of \planneracronym{} in Section \ref{sec:implement}.
Section \ref{exp:Manipulation Tasks} considers a pair of experiments in which simultaneous motion and manipulation tasks are specified. 
Section \ref{exp:Stability-Oriented Scenarios} considers a pair of instability-prone scenarios in which the robot must rely upon the manipulator to remain stable. 
Finally, Section \ref{exp:Time Performance} describes the computational time of \planneracronym{} across these experiments.
These experiments are shown in the attached video.

\subsection{Implementation} \label{sec:implement}


\begin{figure} [!tb]
    \centering
    \begin{subfigure}[b]{0.49\columnwidth}
        \makebox[\textwidth]{%
            \includegraphics[width = 0.85\textwidth]{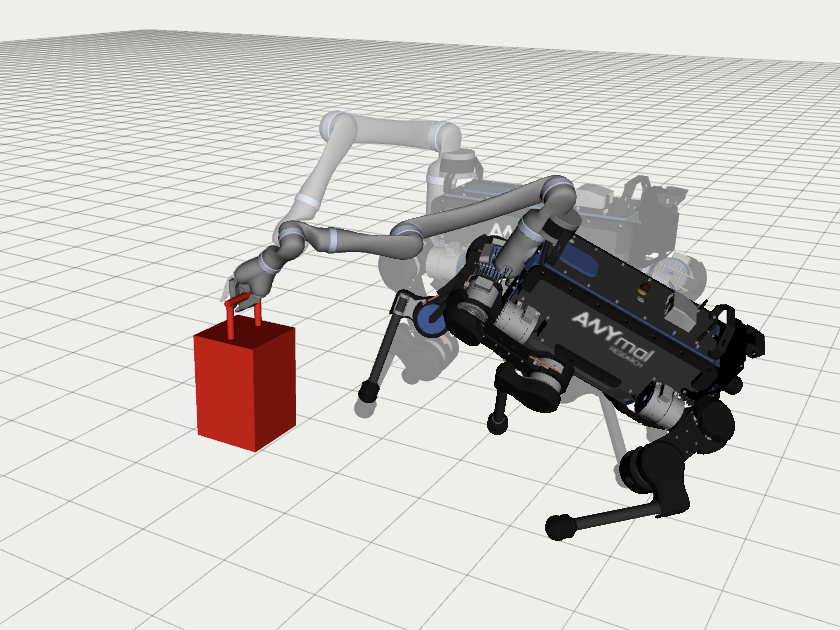}
        }
        \caption{Baseline planner becomes unstable.}
        \label{fig:failed_lift}
    \end{subfigure}
    \hfill
    \begin{subfigure}[b]{0.49\columnwidth}
         \makebox[\textwidth]{%
            \includegraphics[width = 0.85\textwidth]{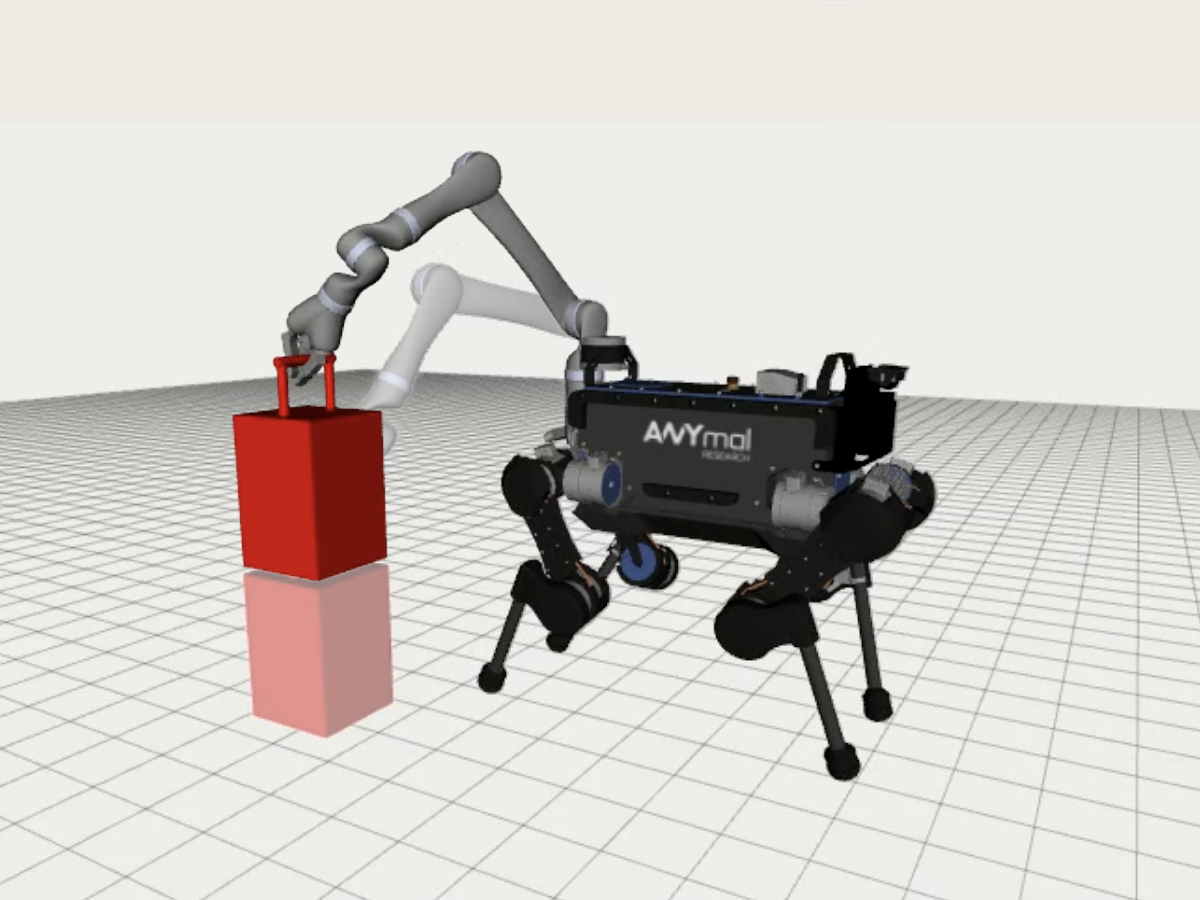}
        }
        \caption{Lifting 3kg using \planneracronym{}.}
        \label{fig:success_lift}
    \end{subfigure}
    \caption{The robot was tasked with lifting a 3kg weight located on the floor. When using the baseline planner, the robot becomes unstable. Using \planneracronym{}, this is done while keeping the robot stable.}
    \label{fig:weight_lift}
\end{figure}

To solve the nonlinear optimization problem in \eqref{original_opt}, this paper relies on sequential quadratic programming (SQP), active-set method. 
A series of experiments are run to show object manipulation and stabilizing behaviours using the Anymal platform \cite{hutter2016anymal} with a 6 DOF Kinova Jaco \cite{campeau2019kinova} robotic arm mounted on its base.
The controller to realize the plans generated by the solution to \eqref{original_opt} uses the robot model in \eqref{equation_of_motion} with hierarchical null space projection \cite{bellicoso2019alma} to control the joint torques of the robot. 
We compare a subset of the experiments with a baseline planner used on Anymal \cite{bellicoso2018dynamic,bellicoso2018optimization}.
This planner does not consider manipulation forces when planning the motion trajectory.
Experiments are run using the Gazebo simulator on a desktop computer with an i5-4590 processor with 16GB of RAM. 
Lastly, because we solve the planning problem in a receding horizon fashion, we use the previously successful plan as an initial guess for the SQP solver.

\subsection{Manipulation Tasks} \label{exp:Manipulation Tasks}

In the first manipulation task, the robot is required to remain stationary while lifting a 3 kg weight using its manipulator as illustrated in Fig. \ref{fig:weight_lift}.
When using \planneracronym{}, the robot in Fig. \ref{fig:success_lift} is able to anticipate the load of the weight and adjust its motion trajectory plan accordingly to remain stable for the duration of the task.
For comparison, we ran the same experiment using the baseline Anymal planner and show the results in Fig. \ref{fig:failed_lift}.
When the manipulation forces at the end-effector are not considered, the plan generated by the baseline planner is unable to allow the robot to lift the weight and the robot becomes unstable and falls over.

\begin{figure} [!tb]
    \centering
    \begin{subfigure}[b]{\columnwidth}
        \includegraphics[width=\textwidth]{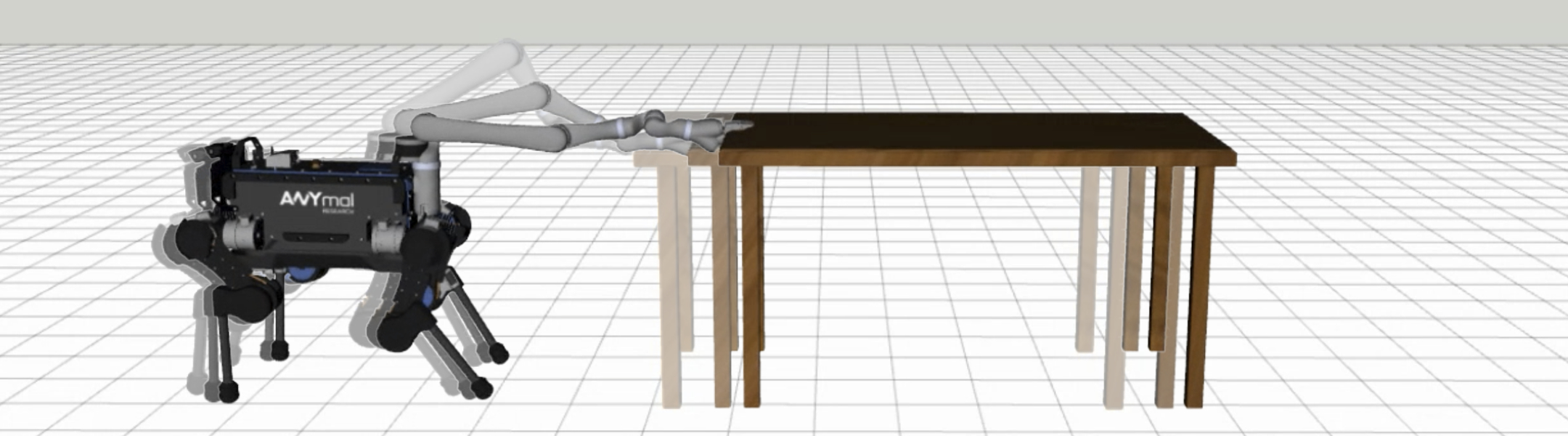}
        \caption{Pushing a table using the baseline planner.}
        \label{fig:table_timeslapse_w/o_STORMMAP}
    \end{subfigure}
    \par
    \begin{subfigure}[b]{\columnwidth}
        \includegraphics[width=\textwidth]{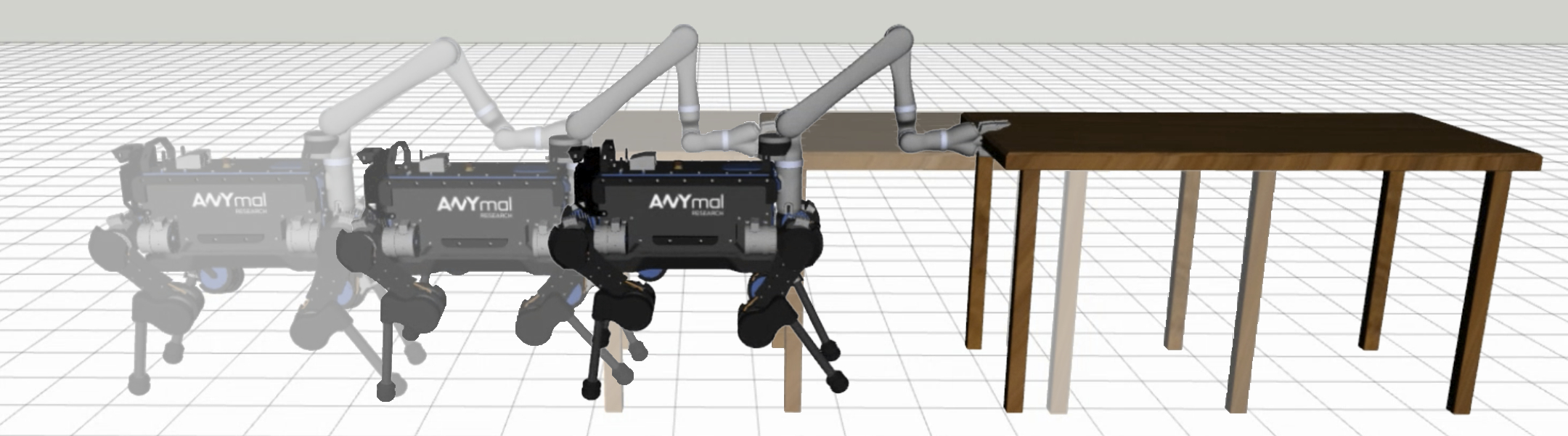}
        \caption{Pushing a table using \planneracronym{}.}
        \label{fig:table_timeslapse_STORMMAP}
    \end{subfigure}
    \par
    \begin{subfigure}[b]{\columnwidth}
        \includegraphics[width=\textwidth]{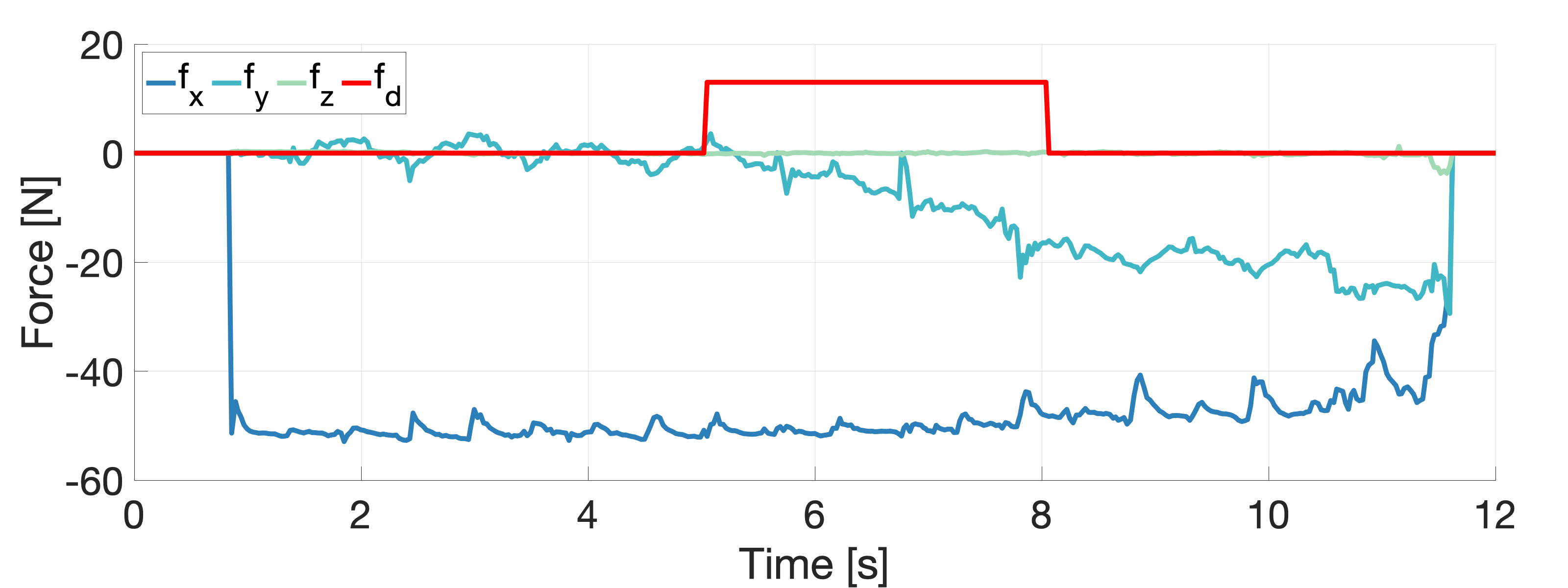}
        \caption{End-effector force while using \planneracronym{} to push the table.}
        \label{fig:table_timeslapse_plot}
    \end{subfigure}
    \caption{Using the baseline planner, Anymal cannot push a table while walking forwards. \planneracronym{} is able to accomplish both the manipulation and motion task.
    To make the task harder a random, unknown disturbance is applied to Anymal perpendicular to the direction of motion.
    Since forces in the \textit{y} direction were not constrained for this task, \planneracronym{} mitigates this disturbance by applying an opposing force to the table. Forces shown are acting on the robot.}
    \label{fig:table_timelapse}
\end{figure}

In the second manipulation task, the robot is required to move a table across a room as illustrated in Fig. \ref{fig:table_timelapse}.
A lateral force of $50N$ is required to push the table, and the coefficient of friction between the hard plastic feet of the robot and the hardwood floor is set at $\mu = 0.3$.
Note that the planner is given this coefficient of friction.
To make this task more difficult we apply a disturbance to the robot's torso.
This disturbance is perpendicular to the robot's motion and is applied 4 seconds after it begins pushing.
The direction and magnitude of the disturbance are unknown to the planner.
Results for the table pushing experiment using \planneracronym{} are depicted in Fig. \ref{fig:table_timeslapse_STORMMAP}. 
From the force graph in Fig. \ref{fig:table_timeslapse_plot} we see a disturbance force of $15N$ applied to the base in the y-direction, shown in red.
During this disturbance, \planneracronym{} modifies the manipulation force plan to counteract this disturbance force, and continues to apply this force to bring the robot back in-line with the table.
Using the baseline Anymal motion planner for this experiment, the robot is able to push the table forward initially, but is unable to maintain an appropriate forward velocity (Fig. \ref{fig:table_timeslapse_w/o_STORMMAP}).
After several seconds of pushing, the manipulator reaches the limit of its reachable space and is unable to push the table further.
Both manipulation experiments are shown in the attached video.

\subsection{Stability-Oriented Scenarios} \label{exp:Stability-Oriented Scenarios}

This subsection describes a pair of experiments, both of which are included in the atached video, wherein the manipulator forces at the end-effector need to be utilized to ensure that the robot is able to complete a given task without falling over.
This is accomplished using the ZMP stability constraint in \eqref{zmp}, in which the force at the manipulator's end-effector is directly tied to the stability of the platform and the friction pyramid constraint in \eqref{friction_pyramid}, which necessitates dynamic feasibility.
We do not compare against the baseline planner for these experiments, as the baseline planner is unable to affect the manipulation forces and thus cannot change them to stabilize the robot.
 
The first experiment considered in this subsection requires that the robot hold onto a railing and remaining stationary while random disturbances are applied to it.
The directions and magnitudes of these disturbances are unknown to the planner.
When applied to this scenario, \planneracronym{} is able to output a manipulation force plan to counteract these disturbances by applying forces of the same approximate magnitude, but in the opposing direction.
This scenario illustrates that \planneracronym{} can use the manipulator quickly to enhance the stability of the robot.

\begin{figure} [!tb]
    \centering
    \begin{subfigure}[b]{\columnwidth}
        \includegraphics[width=\textwidth]{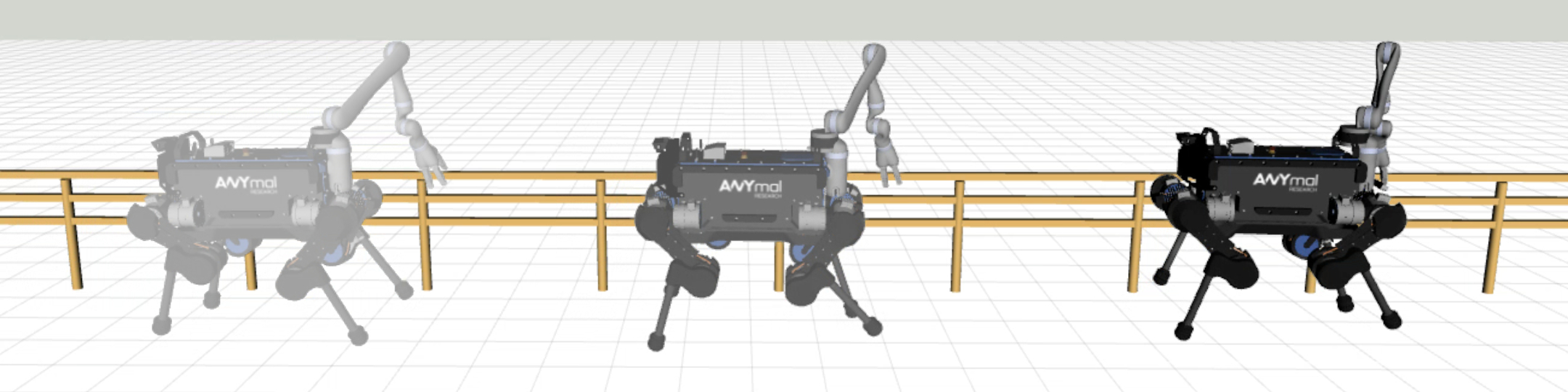}
        \caption{Railing stability with STORMMAP}
        \label{fig:railing_timeslapse_STORMMAP}
    \end{subfigure}
    \par
    \begin{subfigure}[b]{\columnwidth}
        \includegraphics[width=\textwidth]{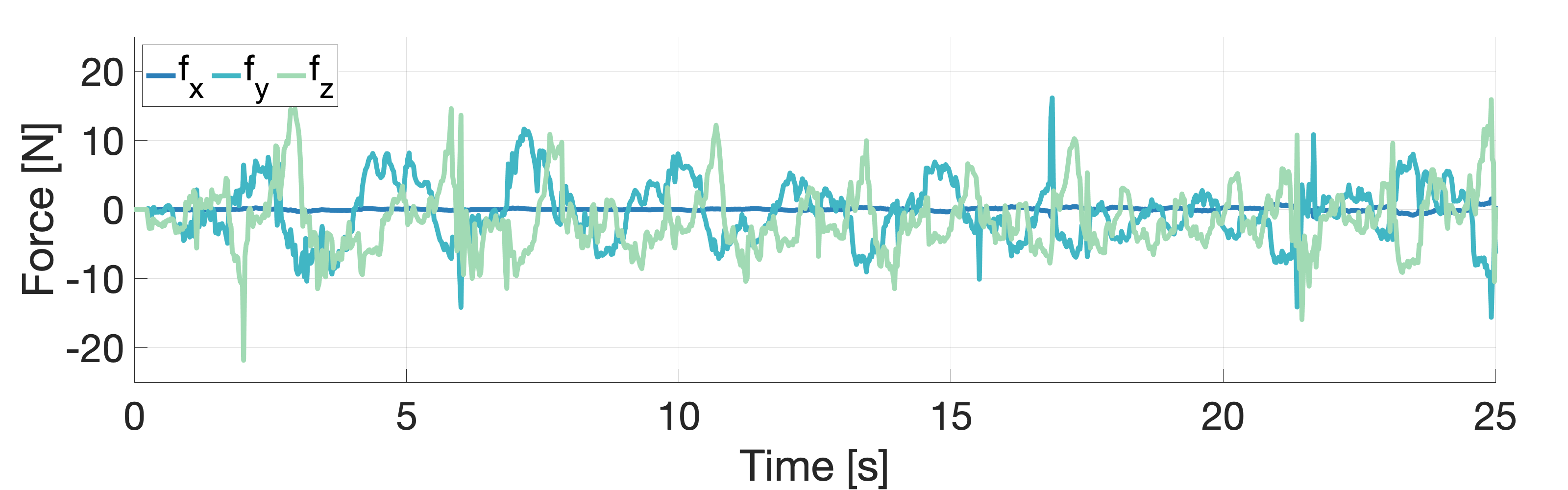}
        \caption{End-Effector force}
        \label{fig:railing_timeslapse_plot}
    \end{subfigure}
    \caption{Anymal walks across a slippery surface while grasping a railing. Only the contact location with the railing is specified by the user while \planneracronym{} optimizes for the forces which limits sway in the body and keep the robot from slipping while it walks forwards. The evolution of the manipulation force plan is plotted. Forces are those acting on the robot.}
    \label{fig:railing_timelapse}
\end{figure}

With the same rail setup, we ran a second experiment in which the robot was required to walk across a slippery floor at a constant desired speed.
During the experiment, the manipulator can slide along the railing and apply forces tangential to but not along the railing (Fig. \ref{fig:railing_timeslapse_STORMMAP}).
As shown in Fig. \ref{fig:railing_timeslapse_plot}, \planneracronym{} optimizes for the manipulation forces at the end-effector in both the \textit{y} and \textit{z} directions. 
The force in the \textit{y} direction keeps the ZMP within the support polygon while reducing the necessary sway of the base.
The force in the \textit{z} direction also changes the position of the ZMP and keeps it within the support polygon.
Additionally, the robot is also able to increase its effective weight by pushing upward on the railing, allowing the robot to exert larger tangential forces at the feet to accelerate forward without having its feet slip on the surface.
No forces are applied in the \textit{x} direction, as we have specified this as a direction of free motion in the constraint in \eqref{freemotion}.

\subsection{Time Performance} \label{exp:Time Performance}

The planning horizon was approximately 0.8 seconds across every test.
When the robot was stationary, the optimization took on average less than 1 ms to compute the motion trajectory and manipulation force plan.
When the robot was walking, the optimization took approximately 7 ms.
This difference is attributed to the fact that only one motion and force spline are needed for the stationary experiments, leading to fewer optimization variables.

The speed of the optimization is such that the \planneracronym{} is able to run in a receding-horizon fashion.
Once a plan is successfully computed, it is stored and used for the robot's controller and the computation of a new plan starts.
\planneracronym{} is, to the best of our knowledge, an order of magnitude faster compared with other planners from the literature based on planning speed \cite{dai2014whole, lin2018offline, lin2019robust, bouyarmane2012humanoid, ponton2016convex}.

\section{DISCUSSION AND CONCLUSION} \label{conclusions}
This paper describes a legged mobile manipulation planner, \planneracronym{}, which can compute plans for both the motion trajectory of the robot as well as the manipulation forces it exerts.
This is done faster than previous dynamic planning strategies and can run in real-time.
By formulating the planner as a nonlinear optimization problem with pre-specified contact locations for the manipulator's end-effector, we can plan dynamically feasible motions and forces while ensuring the stability of the platform.

This was accomplished by using a quintic spline parameterization to reformulate the nonlinear optimization problem used for planning.
We used this formulation in a set of simulated experiments to demonstrate the versatility of \planneracronym{}.
The planner can accomplish manipulation tasks as well as show stability-centric behaviours.
When a manipulation force is specified by the user, the planner can adhere to this desired task force and maintain stability throughout its application.
Likewise, when the contact location, but not the manipulation force, is specified \textit{a-prior}, the planner can optimize for these contact forces to accomplish a desired motion.

In the future we will include a parallel planner to optimize for the contact locations of the manipulator.
This setup would allow the robot to select its own contact locations while hopefully maintaining the computational speed of \planneracronym{}.
Additionally, we will test \planneracronym{} on a physical robot.

\addtolength{\textheight}{-12cm}  




\printbibliography[title={References}]

\end{document}